\documentclass[conference]{IEEEtran}
\IEEEoverridecommandlockouts
\usepackage{cite}
\usepackage{amsmath,amssymb,amsfonts}
\usepackage{algorithmic}
\usepackage{graphicx}
\usepackage{textcomp}
\usepackage{acronym}
\usepackage{xcolor}
\usepackage{bm}
\usepackage{tikz}
\usepackage{subcaption}
\def\BibTeX{{\rm B\kern-.05em{\sc i\kern-.025em b}\kern-.08em
    T\kern-.1667em\lower.7ex\hbox{E}\kern-.125emX}}

\acrodef{GELU}{Gaussian Error Linear Unit}

\newcommand{\x}{\mathbf{x}}
\newcommand{\y}{\mathbf{y}}

\newcommand{\f}{\mathbf{f}}
\newcommand{\F}{\mathbf{F}}

\newcommand{\X}{\mathbf{X}}
\newcommand{\A}{\mathbf{A}}
\newcommand{\Tau}{\mathcal{T}}
\newcommand{\btheta}{\bm{\theta}}

\newcommand\copyrighttext{%
  \footnotesize \textcopyright \the\year{} IEEE. Personal use of this material is permitted. Permission from IEEE must be obtained for all other uses, including reprinting/republishing this material for advertising or promotional purposes, collecting new collected works for resale or redistribution to servers or lists, or reuse of any copyrighted component of this work in other works.}

\newcommand\copyrightnotice{%
\begin{tikzpicture}[remember picture,overlay]
\node[anchor=south,yshift=10pt] at (current page.south) {\fbox{\parbox{\dimexpr0.75\textwidth-\fboxsep-\fboxrule\relax}{\copyrighttext}}};
\end{tikzpicture}%
}

\begin{document}
% Definitions of the acronyms
\acrodef{AI}{Artificial Intelligence}
\acrodef{RBC}{Rayleigh-Bénard Convection}
\acrodef{NSSE}{Normalized Sum of Squared Errors}
\acrodef{DMD}{Dynamic Mode Decomposition}
\acrodef{PDE}{partial differential equation}
\acrodef{DMD}{Dynamic Mode Decomposition}
\acrodef{LRAN}{Linear Recurrent Autoencoder Network}
\acrodef{DNS}{Direct Numerical Simulation}

\title{Koopman-Based Surrogate Modelling of Turbulent Rayleigh-Bénard Convection
\thanks{The authors acknowledge financial support by the project “SAIL: SustAInable Life-cycle 
of Intelligent Socio-Technical Systems” (Grant ID NW21-059A), which is funded by the 
program “Netzwerke 2021” of the Ministry of Culture and Science of the State of 
North Rhine Westphalia, Germany.}
}

\author{\IEEEauthorblockN{1\textsuperscript{st} Thorben Markmann}
\IEEEauthorblockA{\textit{Technische Fakultät} \\
\textit{Bielefeld University}\\
Bielefeld, Germany \\
tmarkmann@techfak.uni-bielefeld.de}
\and
\IEEEauthorblockN{2\textsuperscript{nd} Michiel Straat}
\IEEEauthorblockA{\textit{Technische Fakultät} \\
\textit{Bielefeld University}\\
Bielefeld, Germany \\
mstraat@techfak.uni-bielefeld.de}
\and
\IEEEauthorblockN{3\textsuperscript{rd} Barbara Hammer}
\IEEEauthorblockA{\textit{Technische Fakultät} \\
\textit{Bielefeld University}\\
Bielefeld, Germany \\
bhammer@techfak.uni-bielefeld.de}
}

\maketitle
\copyrightnotice

\begin{abstract}
Several related works have introduced Koopman-based Machine Learning architectures as a surrogate model for dynamical systems. These architectures aim to learn non-linear measurements (also known as observables) of the system's state that evolve by a linear operator and are, therefore, amenable to model-based linear control techniques. So far, mainly simple systems have been targeted, and Koopman architectures as reduced-order models for more complex dynamics have not been fully explored. Hence, we use a Koopman-inspired architecture called the Linear Recurrent Autoencoder Network (LRAN) for learning reduced-order dynamics in convection flows of a Rayleigh Bénard Convection (RBC) system at different amounts of turbulence. The data is obtained from direct numerical simulations of the RBC system.
A traditional fluid dynamics method, the Kernel Dynamic Mode Decomposition (KDMD), is used to compare the LRAN.
For both methods, we performed hyperparameter sweeps to identify optimal settings. We used a Normalized Sum of Square Error measure for the quantitative evaluation of the models, and we also studied the model predictions qualitatively.
We obtained more accurate predictions with the LRAN than with KDMD in the most turbulent setting. We conjecture that this is due to the LRAN's flexibility in learning complicated observables from data, thereby serving as a viable surrogate model for the main structure of fluid dynamics in turbulent convection settings. In contrast, KDMD was more effective in lower turbulence settings due to the repetitiveness of the convection flow.
The feasibility of Koopman-based surrogate models for turbulent fluid flows opens possibilities for efficient model-based control techniques useful in a variety of industrial settings.

\end{abstract}

% Same keywords as used for APPIS, maybe extend...

\section{Introduction}
Modern dynamical settings are characterized by the availability of large amounts of data that are obtained from measurements by increasingly prevalent sensors or sophisticated computer simulations.
Moreover, the growing availability of fast computing and storage capacity creates many possibilities for applying modern Machine Learning techniques to dynamical systems \cite{brunton_data_2022,pandey2022}.

Simultaneously, urgent challenges in disciplines such as climate science and epidemiology call for efficient tools for modeling, prediction, and control of large-scale, complex dynamical systems.
Accurate models based on first principles for these complex dynamical systems do often not exist. Moreover, direct numerical simulations of large-scale systems can quickly become computationally infeasible.

Although the spatial resolution of measurements has been increasing with technological and computational developments, the underlying dynamical structure is frequently located on a low-dimensional manifold \cite{brunton_data_2022}.
Additionally, the interpretation of dynamical systems is improved by decomposing the complex dynamics in simpler constituents \cite{chatterjee2000,kutz_dmd_2016}.
Finding these coherent structures from the measurements of the system is critical for efficient modeling and the prediction of the system's future state. Models for representing dynamics in a few main components are known in the literature as \textit{reduced-order models} \cite{pandey2022,pawar_deep_2019,Brunton_2016}.

A prominent method for reduced order modeling
%for finding coherent structures
of dynamical systems is the \ac{DMD}, which was originally introduced in the fluid dynamics field \cite{schmid_2010}. The \ac{DMD} represents the dynamics as a linear combination of the main \textit{dynamic modes}. The modes evolve and can be used to make predictions of the system's future state. Extended \ac{DMD} \cite{williams_2015_edmd} is an extension aiming at improving the approximation quality by introducing a fixed dictionary of non-linear measurements of the system. The kernel trick was used for efficient computation in the Kernel \ac{DMD} method \cite{Williams2015_kdmd}.

Recent works have focused on identifying \textit{intrinsic coordinates} of a dynamical system from its measurements by using \textit{Koopman theory}, which has its origins in \cite{koopman_1931}.
That work established that any non-linear dynamical system has a linear counterpart in a potentially infinite number of non-linear measurement functions or \textit{observables} of the system's state. The operator evolving the observables linearly is called the \textit{Koopman operator}.
Coherent structures of the dynamics called \textit{Koopman modes} (similar to the dynamic modes in the \ac{DMD}) can be extracted from the Koopman operator. In \cite{rowley_spectral_2009}, it is therefore argued that \ac{DMD} methods are essentially algorithms for approximating Koopman modes.

The use of Machine Learning methods in approximating Koopman representations of dynamical systems has seen increased interest \cite{Brunton_modernKoopman_2022}. This is mainly due to the ease of handling linear dynamical systems and control.
In these Machine Learning methods, a finite number of invertible observables that evolve approximately linearly is learned from measurements using encoder-decoder neural networks \cite{lusch_deep_2018,otto2019,bonnert_estimating_2020,takeishi_learning_2017,yeung_learning_2019,klie_2020,morton_2018,Brunton_2016}.

In particular, the \ac{LRAN} architecture, introduced in \cite{otto2019}, uses an autoencoder to learn observables of the system states. The time-stepping in the observable space is linear by a simple matrix multiplication. Hence, the architecture approximates the Koopman operator by a matrix multiplication.
The cost function consists of terms that steer the weights of the architecture to realize an autoencoder with both good reconstruction quality and linear dynamics in the latent space. The method was demonstrated on the duffing, cylinder wake and Kuramoto-Sivashinsky dynamical systems \cite{otto2019}.
In recent years, these machine Koopman-based network architectures have been explored further in diverse dynamical systems relevant to real-world problems \cite{schulze_2022,zhao_2023,klie_2020,morton_2018}.

% Motivation for our work.
In this work, we study the effectiveness of an architecture inspired by the \ac{LRAN} for learning the dynamics of thermal convection processes in fluids modeled by \ac{RBC}.
Efficient modeling of thermal convection is an important research area due to its ubiquity in nature and relevance to many industrial settings \cite{pandey_2018}.
It is frequently required to control these systems to a desired state, e.g. suppressing or enhancing the convective heat exchange \cite{beintema_2020}.
In this case, model-based control techniques are often more efficient than model-free techniques. As an initial step towards model-based control, our current work focuses on modeling the uncontrolled \ac{RBC} dynamics from the simulation data.
%In addition, a Koopman-based surrogate model is beneficial in comparison to non-linear approaches for the relative ease of incorporating control in the linear dynamics. 

Several works in the literature focus on surrogate models for \ac{RBC}. For instance, in \cite{pandey2022}, an autoencoder is trained initially to reduce the dimension of the convective fields obtained from the simulations of \ac{RBC}. Afterward, a GRU sequence model is trained and compared to a reservoir network for the modeling of the latent space dynamics. In contrast, in our work, we aim to find state encodings in which the dynamics are linear.

The paper is structured as follows: In Sec.~\ref{sec:methodology}, we provide an overview of the Rayleigh-Bénard system and the direct numerical simulation we used. We then provide a brief formal description of Koopman theory, after which we introduce the kernelized version of the DMD method and the LRAN as surrogate models for the dynamics. In Sec.~\ref{sec:experiment_settings}, we detail the experiment settings we used for evaluating the models in different convection situations. Subsequently, in Sec.~\ref{sec:results} we give the results and discuss them in Sec.~\ref{sec:discussion}. We end with a summary and an outlook for future work in Sec.~\ref{sec:conclusion}.

%In \cite{beintema_2020}, the authors focus on reducing the convective heat transport in \ac{RBC}  by designing a reinforcement learning based controller work with sub-sampled domain of the original system 

\section{Methodology} \label{sec:methodology}

\subsection{Rayleigh-Bénard Convection}
Rayleigh-Bénard Convection describes a layer of fluid heated at the bottom with the following partial differential equation \cite{mortensen_rbc}:
% We consider dynamical systems that that can be described by a \ac{PDE}
% \begin{equation} \label{eq:partial_diff}
%     \mathbf{u}_t = \mathbf{N}(\mathbf{u}, \mathbf{u}_x, \mathbf{u}_{xx},...,x,t;\bm{\beta})\, ,
% \end{equation}
% where $\mathbf{u}$ is the system's state, $\bm{\beta}$ are the \ac{PDE} parameters, $\mathbf{N}$ is a nonlinear evolution operator and $x$ and $t$ is the spatial and temporal variable, respectively.
%Although the methodology is applicable to a wider array of systems \eqref{eq:partial_diff}, we build and apply the methods on the two-dimensional \ac{RBC} system from fluid dynamics.

\begin{flalign} \label{eq:RBC_eqs}
\begin{split}
& \frac{\partial \bm{u}}{\partial t} + (\bm{u} \cdot \nabla) \bm{u} = -\nabla p + \sqrt{\frac{Pr}{Ra}} \nabla^2 \bm{u} + T \bm{j}\,, \\
& \frac{\partial T}{\partial t} + \bm{u} \cdot \nabla T = \frac{1}{\sqrt{Ra Pr}} \nabla^2 T\,, \\
& \nabla \cdot \bm{u} = 0\,,
\end{split}
\end{flalign}
where $\bm{u}=u_x \bm{i} + u_y \bm{j}$ is the velocity vector field, $\bm{i}$ and $\bm{j}$ are unity vectors in the $x$ and $y$ axis directions, $p$ is the pressure field and $T$ is the temperature field. Note that the equations have been non-dimensionalized following \cite{pandey_2018} and that the pressure field can be eliminated from the dynamics \cite{mortensen_rbc}. Consequently, the temperature and velocity fields comprise the system's state.

The equations have two parameters: the Prandtl number $Pr$ and the Rayleigh number $Ra$.
The Prandtl number $Pr$ describes the fluid's ratio of kinematic viscosity to thermal diffusivity.
The Rayleigh number $Ra$ determines the thermal driving in convection \cite{pandey_2018} and is proportional to the temperature gradient between the bottom and top layers of the cell.
In this work, we keep the Prandtl number fixed at $0.7$ and perform our experiments for increasing Rayleigh numbers, which results in higher degrees of convective turbulence in the fluid.
The convective heat transfer is expressed by the local-convective heat flux \cite{pandey2022}:
\begin{align} \label{eq:convective}
\begin{split}
    q(\x, t) &= u_y(\x, t) \theta(\x, t)\, \\ 
    \text{ with } \theta(\x, t) &= T(\x, t) - \langle T \rangle_{x, y, t}\,,
\end{split}
\end{align}
where $\theta$ is the temperature relative to the average temperature over space $\langle T \rangle_{x, y, t}$ at time $t$.
A principal quantity in the \ac{RBC} system is the \textit{Nusselt number}. It is defined as the ratio of convective to conductive heat transfer. We consider the Nusselt number over the entire convection field at time $t$, which is easily computed by a spatial average over the field \cite{pandey2022}, i.e.:
\begin{equation}
    Nu(t) = \frac{\langle q(t) \rangle_{x, y}}{\kappa (T_b - T_t) / H},
\end{equation}
where $\kappa$ is the thermal diffusivity, $T_b$ and $T_t$ are the boundary temperatures of the bottom and top layers, respectively, and $H$ is half of the domain height.

An \ac{RBC} episode starts with heat transfer only by conduction. For Rayleigh numbers above $Ra=1708$, thermal convection is induced by the dominance of gravitational forces. These flows become more turbulent for increasing $Ra$. %The position of the convective cells is not predictable from the initial conditions and small deviations can cause large shifts in the cells.

\subsection{Direct Numerical Simulation (DNS)}
\label{sec:dns}
\begin{figure*}
    \centering
    \begin{subfigure}[b]{0.35\textwidth}
        \centering
        \includegraphics[width=\textwidth]{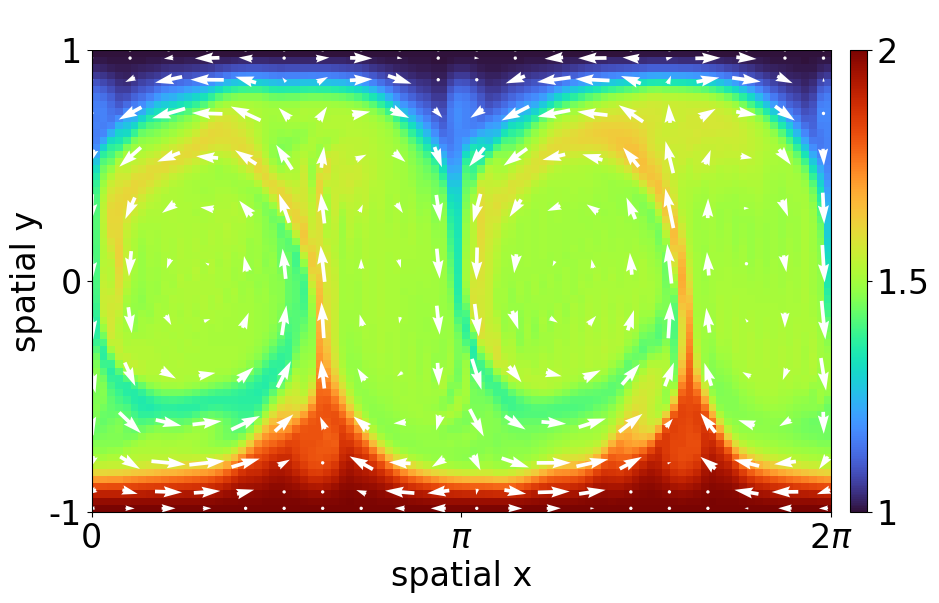}
        %\caption{}
        %\label{fig:example_state}
    \end{subfigure}
    \begin{subfigure}[b]{0.35\textwidth}
        \centering
        \includegraphics[width=\textwidth]{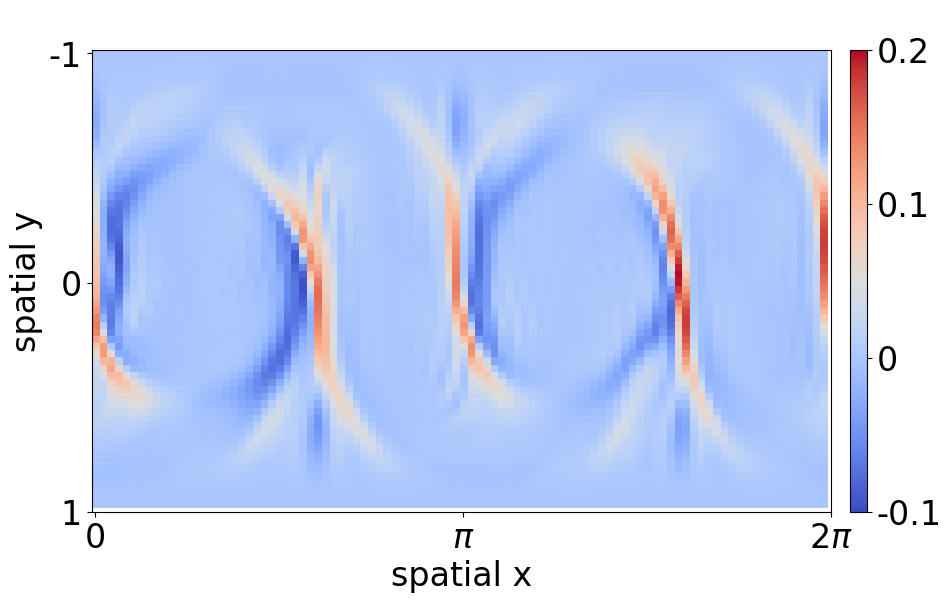}
        %\caption{}
        %\label{fig:example_conv}
    \end{subfigure}
    \caption{\textit{Left:} Rayleigh Benard Convection state. Color is temperature, arrows fluid velocity. \textit{Right:} Corresponding local convective field. The state results from a Direct Numerical Simulation for Rayleigh number $Ra=2e6$.}
    \label{fig:examples}
\end{figure*}
\begin{table}
\centering
\begin{tabular}{||c c||} 
 \hline
 Parameter & Value \\ 
 \hline\hline
 Domain Size & $2\pi \times 2$\\ 
 Galerkin Modes & $96 \times 64$\\
 $Ra$ & \{ $1e5$, $1e6$, $2e6$, $5e6$ \} \\
 $Pr$ & 0.7\\
 ($T_{\mathrm{t}}, T_{\mathrm{b}})$ & (1,2)\\
 $\Delta t$ & 0.025\\
 Episode Length & 500\\
 Cook Time & 100\\
 \hline
\end{tabular}
\caption{System Parameters used in the simulation for Rayleigh-Bénard Convection rollouts.}
\label{table:dns}
\end{table}
Throughout this work, ground truth data for rollouts of the Rayleigh-Bénard Convection have been obtained through a \ac{DNS}. The equations in \eqref{eq:RBC_eqs} are solved numerically in two dimensions using the spectral Galerkin method. We utilize the open-source framework Shenfun \cite{mortensen_joss, shenfun}, which provides a solver for the Rayleigh Benard Convection. For our \ac{DNS}, we use a setup similar to that of \cite{vignon_2023} with system parameters summarized in Table \ref{table:dns}. This also includes no-slip boundary conditions at the bottom and top and a periodic boundary condition in the horizontal direction of the fields. Starting from an initial condition given by random noise on top of the diffusive equilibrium, the system is sampled at a rate of $\Delta t=0.025$. Since we focus on states in the convective phase of the Rayleigh-Bénard Convection, simulation data is recorded after the first 100 time units. Moreover, as our focus is on the convective properties of RBC, we only consider the local convective field given by Equation \eqref{eq:convective} instead of the whole system state consisting of a temperature and velocity field. Examples for both local convective and whole state are shown in Figure \ref{fig:examples}.

\subsection{Koopman theory for dynamical systems}
% Discuss the most important ideas regarding Koopman theory for dynamical systems and introduce the main equations.
Here we give a brief overview of Koopman analysis based on \cite{Brunton_modernKoopman_2022}, which can be consulted for more details.

The \textit{flow-map operator} $\F^{\Delta t}(\cdot)$ applied to a dynamical system such as \eqref{eq:RBC_eqs} is defined as:
\begin{equation} \label{eq:flow_map}
\x_{k+1} = \F^{\Delta t}(\x_k) = \x_k + \int_{k\Delta t}^{(k+1)\Delta t} \f(\x(\tau)) d\tau\,,
\end{equation}
which yields discrete samples $\x_k = \x(k \Delta t)$.
The system \eqref{eq:flow_map} can be analysed in terms of \textit{observables} $g_j(\x)$, $g_j : \mathbb{R}^N \to \mathbb{R}$.
%which are elements of a potentially infinite-dimensional Hilbert space of functions of the system's state. In \cite{koopman_1931}, the existence of a linear operator $\mathcal{K}^{\Delta t}$ was proven that evolves the observables forward in time:
The linear Koopman operator \cite{koopman_1931} evolves the observables forward in time:
\begin{equation} \label{eq:Koperator}
\mathcal{K}^{\Delta t} g(\x_k) = g(\F^{\Delta t}(\x_k)) = g(\x_{k+1})\,.
\end{equation}
Hence the operator $\mathcal{K}^{\Delta t}$ and the flow-map $\F^{\Delta t}(\cdot)$ operator are counterparts for evolving vectors in their respective spaces.

Koopman eigenfunctions are special observables that evolve according to:
% \begin{equation}
% \mathcal{S}=\{ \phi_i(\x) \}_{i=1}^p, \text{ with } \phi_i : \mathbb{R}^N \to \mathbb{C}\,.
% \end{equation}
\begin{equation}\label{eq:apply_operator}
\mathcal{K}^{\Delta t} \phi_i(\x) = \lambda_i \phi_i(\x)\,,  
\end{equation}
where $\lambda_i \in \mathbb{C}$ is the eigenvalue corresponding to eigenfunction $\phi_i(\x)$.
The aim is to find a $p$-dimensional subspace of the most dominant eigenfunctions and identify using a Machine Learning architecture $n \geq p$ number of observables $g(\x)$ that span that subspace and that can also be inverted to the original states $\x \in \mathbb{R}^N$. The matrix $K \in \mathbb{R}^{n \times n}$ is a finite-dimensional approximation of the Koopman operator.
\textit{Koopman modes} $\bm{\xi}_k \in \mathbb{C}^N$ are vectors in the state-space that linearly combine with the eigenfunctions to reconstruct the state and the flow-map operator:
\begin{equation} \label{eq:koopdecomp}
    \x = \sum_{k=1}^p \bm{\xi}_k \phi_k(\x)\,, \quad \mathbf{F}(\x) = \sum_{k=1}^p \lambda_k \bm{\xi}_k \phi_k(\x)\,.
\end{equation}

\subsection{Dynamic Mode Decomposition}
When measurements are available of a dynamical system, the system's Koopman mode decomposition \eqref{eq:koopdecomp} can be approximated by \ac{DMD} methods \cite{Williams2015_kdmd,kutz_dmd_2016}.
The input to \ac{DMD} is a snapshot matrix $\X \in \mathbb{R}^{N \times (m-1)}$ with
in its columns $m-1$ snapshots $\x(k\Delta t)$ of the system and another matrix $\X' \in \mathbb{R}^{N \times (m-1)}$ with the snapshots shifted $\Delta t$ forward in time.
The matrix of a locally linear approximation $\X' \approx \mathbf{A} \X$ to the dynamics is computed using a least square fit: $\A = \X' \X^\dagger$.
% \begin{equation} \label{eq:linearDMD}
% \X' \approx \mathbf{A} \X\,,
% \end{equation}
Note that \ac{DMD} methods are particularly efficient in situations
where the number of snapshots $m$ is significantly lower than their dimension $N$.
The eigenvectors of $\A$ are called \textit{dynamic modes} and using the corresponding eigenvalues,
the \ac{DMD} solution to the fitted linear dynamical system can be written as: 
\begin{equation} \label{eq:DMDFormulation}
   \x(t) \approx \sum_{k=1}^r \bm{\xi}_k \exp(\omega_k t) b_k\,,
\end{equation}
where $\omega_k = \ln(\lambda_k) / \Delta t$ and $b_k$ is the initial amplitude of each mode \cite{kutz_dmd_2016}.
This version of \ac{DMD} can be understood as an approximation of the Koopman operator using only linear observables. It is, therefore, quite limited in the dynamics that can be represented.

% Describe the extension of kernel mode decomposition.
To approximate the Koopman operator more accurately, non-linear observables of the system's state should be introduced.
Generally, either the extended DMD or the kernel DMD is used depending on the ratio of snapshots $m$ to observables $n$.
In our case where $m \ll n$, the \textit{Kernel Dynamic Mode Decomposition} (KDMD) is the most efficient \cite{kutz_dmd_2016,Williams2015_kdmd}.
The kernel method relies on computing inner products in the observable space. In particular, it is required to compute
\begin{equation} \label{eq:kernelDMD}
    \hat{\bm{G}}^{(ij)} = \bm{g}(\x_j)^T \bm{g}(\x_i)\,, \quad \hat{\bm{A}}^{(ij)} = \bm{g}(\x_j)^T \bm{g}(\y_i)\, ,
\end{equation}
which can be done efficiently using the kernel trick \cite{Williams2015_kdmd}. Note that $\bm{g}(\x_i) \in \mathbb{R}^n$ denotes the vector of all observables evaluated for the state $\x_i$ and $\bm{y}_i = \F^{\Delta t}(\x_i)$. The chosen kernel function $f : \mathbb{R}^N \times \mathbb{R}^N \to \mathbb{R}$ is responsible for implicitly computing the inner products in the observable space:
\begin{equation} \label{eq:kernelDMD_f}
    \hat{\bm{G}}^{(ij)} = f(\x_i, \x_j)\,, \quad \hat{\bm{A}}^{(ij)} = f(\y_i,\x_j)\, ,
\end{equation}
which brings the time complexity down to $\mathcal{O}(m^2 N)$ instead of the $\mathcal{O}(m^2 n)$ required for the computation via explicit transformation as in Eq.~\eqref{eq:kernelDMD}.
Given the matrices in \eqref{eq:kernelDMD_f}, it is possible to approximate the Koopman eigenvalues $\lambda_k$, modes $\bm{\xi}_k$ and eigenfunctions $\phi_k(\x)$ to be used in the Koopman mode decomposition
\eqref{eq:koopdecomp}. We refer to \cite{Williams2015_kdmd} for more details and derivations.

\subsection{Linear Recurrent Autoencoder Network (LRAN)}
\begin{figure}
    \centering
    \includegraphics[width=0.48\textwidth]{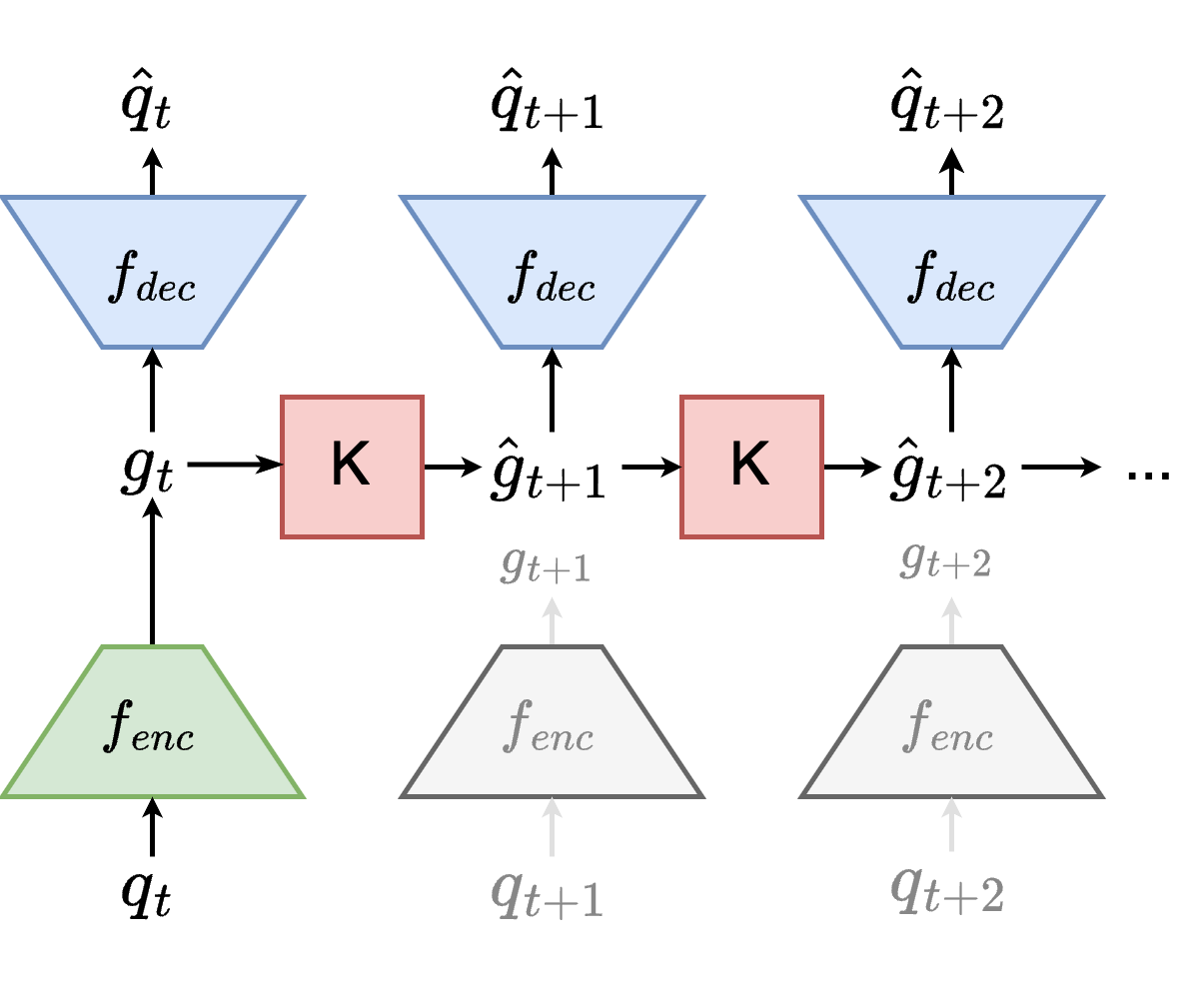}
    \caption{Illustration of the Linear Recurrent Autoencoder Network architecture. Starting from the bottom left the encoder $f_{enc, \theta}$ compresses snapshot $q_t$ to observables $g$. The matrix K evolves the observable forward in time. The decoder $f_{dec, \theta}$ lifts the observables back into the original space to obtain predictions $\hat q$. Grayed-out snapshot compressions $q_{t+1}$ and subsequent are only used during training.}
    \label{fig:lran}
\end{figure}

% Start of the part that discusses learning observables and Koopman from data

Machine Learning architectures were introduced that employ an encoder-decoder structure with a linear dynamical model in the latent space \cite{otto2019,bonnert_estimating_2020,klie_2020}. One of those structures is the \ac{LRAN} introduced in \cite{otto2019}, which is shown in Figure \ref{fig:lran}.
The encoder's learning objective is to find a good manifold in the latent space from which the decoder can reconstruct the convection fields $q$. In addition, these latent variables should act like Koopman observables, i.e., evolve linearly in time using the matrix $K$.
We implement the encoder-decoder structure as a Convolutional Autoencoder (CAE) architecture with five layers for encoder and decoder each (Appendix \ref{app:cae}). CAEs are especially suited because, unlike dense neural networks, the sparsely connected convolutional layer can handle spatially high-dimensional data and capture local dependencies in the data. We use GELU as an activation function and utilize filters of size (5x5) for the 2D convolutional layers.

The whole architecture is trained on snapshot sequences $(q_t, q_{t+1}, ..., q_{t+\Tau-1})$ of varying sequence length $\Tau$. The optimization goals are expressed by the loss function which reads for one example sequence as follows \cite{otto2019}:
\begin{small}
\begin{equation}
     \mathcal{L} = 
      \sum_{\tau=0}^{\Tau-1} \frac{\delta^\tau}{N_1} \frac{\lVert \hat{q}_{t+\tau} - q_{t+\tau}\rVert^2}{\lVert q_{t + \tau} \rVert^2 + \epsilon_1}  
     + \beta \sum_{\tau=1}^{\Tau-1} \frac{\delta^{\tau-1}}{N_2} \frac{\lVert \hat{g}_{t+\tau} - g_{t+\tau}\rVert^2}{\lVert g_{t + \tau} \rVert^2 + \epsilon_2} \,,
\end{equation}
\end{small}
which must be minimized with respect to the encoder, decoder, and linear system parameters $\btheta_{enc}$, $\btheta_{dec}$, and $K$ (see Fig.~\ref{fig:lran} for the flow of information and the meaning of the symbols). The minimization of all parameters is done simultaneously by Adaptive Moment Estimation (ADAM) \cite{kingma_2015} using gradients accumulated on mini-batches of sequences of the training set.
The loss contains the normalized reconstruction error and the normalized hidden error over the predicted sequence. The hyperparameter $\beta$ defines the balance between these two errors and $0 < \delta \leq 1$ implements a decay of importance over time. $N_1$ and $N_2$ are normalization factors that realize a weighted mean.

\subsection{Evaluation}
\label{sec:methods:evaluation}
% Should we come up with a measure to summarize the accuracy over the whole sequence? Average of mean square errors?
We evaluate the model predictions by comparison with the data obtained from the \ac{DNS}. To quantify the error, we use the \ac{NSSE}, which is defined as:
\begin{equation}
    NSSE = \frac{||q - \hat{q}||^2}{||q||^2}\,,
\end{equation}
where $q$ is the ground truth convection field and $\hat{q}$ the predicted one.

\section{Experiment settings} \label{sec:experiment_settings}
\begin{table}
    \begin{subtable}{0.48\textwidth}
        \centering
        \begin{tabular}{||c c||} 
            \hline
            Parameter & Value \\ 
            \hline\hline
            Rayleigh Number & $\in$ \{ 1e5, 1e6, 2e6, 5e6 \} \\
            \hline
            Latent Dimension &  in [16, 1024] \\ 
            $\delta$ &  in [0.9, 1.0] \\
            $\beta$ &  in [0.0, 10.0] \\
            Sequence Length & in [2, 30] \\
            Learning Rate & $\in$ \{ $10^{-4}$, $10^{-5}$ \} \\
            Episode Index & $\in$ \{ 0, 1, 2, 3, 4 \} \\
            \hline
            Epochs & 50 \\
            Batch Size & 32 \\
            \hline
        \end{tabular}
        \caption{LRAN hyperparameter}
        \label{table:sweep:lran}
    \end{subtable}
    \newline
    \vspace*{0.5 cm}
    \newline
    \begin{subtable}{0.48\textwidth}
        \centering
        \begin{tabular}{||c c||} 
            \hline
            Parameter & Value \\ 
            \hline\hline
            $\sigma$ & $\in$ \{ 1, 2, 4, 6 \} \\ 
            Snapshot Size & $\in$ \{ 5, 10, 30, 40, 60, 80, 100, 150 \} \\
            \hline
        \end{tabular}
        \caption{KDMD hyperparameter}
        \label{table:sweep:kdmd}
    \end{subtable}
    \caption{Hyperparameter search ranges for Experiment 1 and 2}
    \label{table:sweep}
\end{table}
First, we conduct hyperparameter tuning for LRAN and KDMD in experiments 1 and 2, respectively. For both methods, multiple parameter searches are performed separately for various Rayleigh numbers to find suitable parameter configurations under increasing turbulence in the RBC system. Afterward, we compare both methods in Experiment 3.

\subsection{Train and test data}
The DNS from Section \ref{sec:dns} is used to generate four different datasets at $Ra \in \{ 1e5, 1e6, 2e6, 5e6 \}$. Each dataset contains system states from 25 simulation episodes, starting from different initial conditions. Even though the DNS runs at a smaller $\Delta t=0.025$, we only consider whole timesteps for an episode length of $T=500$. The episodes are split into a train and a test phase, transitioning at $t=470$. Therefore, the test data consists of a single prediction window $( q_{470}, q_{471},..., q_{499} )$ of length $\Tau_{\mathrm{test}}=30$. Depending on the model, we choose the training data from the first 470 timesteps.

\subsection{Experiment 1: Finding the right LRAN architecture}
For increasing $Ra$ a different set of hyperparameters may work out better. Therefore, we conduct hyperparameter sweeps to find suitable parameter configurations for each Rayleigh number separately.
For the LRAN, the tunable hyperparameters are the length of the train sequences $\Tau$, the number of observables, the decaying weight $\delta$, and the hidden loss weight $\beta$ of the loss function. The ranges of possible values for these parameters are depicted in Table \ref{table:sweep:lran}. During the search, a random configuration is determined and used to train the LRAN on a single episode. Depending on the sequence length, the data used for training consists of $470-\Tau$ overlapping sequences $ ( q_{t+0},..., q_{t+\tau-1} )$. These are further split into a training and validation set, containing randomly selected 80\% and 20\% of sequences, respectively. Instead of fixing the number of training epochs, early stopping is performed if the validation loss plateaus for more than five epochs. Afterward, the model is evaluated on the test window using the NSSE metric from Section \ref{sec:methods:evaluation}. Overall, around 100 training runs are performed for each $Ra$ to guarantee a sufficient covering of the hyperparameter landscape.

\subsection{Experiment 2: Tuning parameters of KDMD}
Analogously, we conduct a hyperparameter search for KDMD for the same four Rayleigh Numbers. Here, the main parameters are the snapshot matrix size $\Tau$ and the width of the Gaussian kernel sigma. Parameter ranges are shown in Table \ref{table:sweep:kdmd}. KDMD is evaluated using the NSSE on the same episodes and test windows as the LRAN. Accordingly, the snapshot matrix is defined by $\bm{X} = ( q_{470-\tau}, ..., q_{469} ) $.
We use a grid search to test each possible distinct configuration, resulting in 32 runs for each of the four Rayleigh Numbers.

\subsection{Experiment 3: Comparing LRAN and KDMD}
After searching for reasonable hyperparameter configurations for both methods at different levels of turbulence, we compare both methods in experiment 3. The optimal configurations determined from the previous experiments are used to evaluate both methods for the four datasets with varying $Ra$.

\section{Results} \label{sec:results}
% Show the results of the experiments.
\subsection{Experiment 1}
% Sweep Results
\begin{figure}
    \centering
    \begin{subfigure}{0.24\textwidth}
        \centering
        \includegraphics[width=\textwidth]{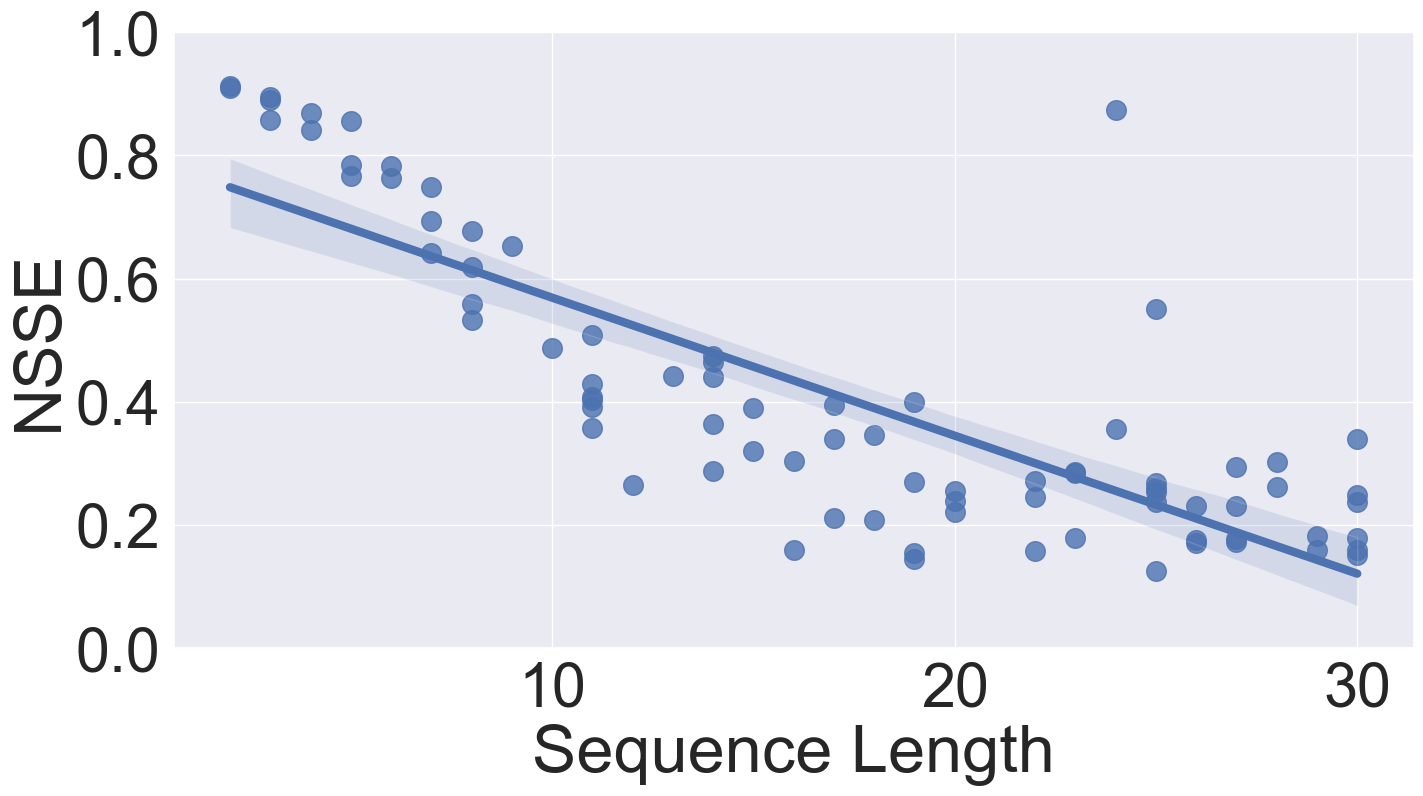}
        \caption{}
        \label{fig:results:sweep:ra1e6:seq}
    \end{subfigure}
    %\begin{subfigure}{0.24\textwidth}
    %    \centering
    %    \includegraphics[width=\textwidth]{figures/sweep/sweep_ra1e6/ra1e6_latent.png}
    %    \caption{}
    %    \label{fig:results:sweep:ra1e6:latent}
    %\end{subfigure}
    %\begin{subfigure}{0.24\textwidth}
    %    \centering
    %    \includegraphics[width=\textwidth]{figures/sweep/sweep_ra1e6/ra1e6_delta.png}
    %    \caption{}
    %    \label{fig:results:sweep:ra1e6:delta}
    %\end{subfigure}
    \begin{subfigure}{0.24\textwidth}
        \centering
        \includegraphics[width=\textwidth]{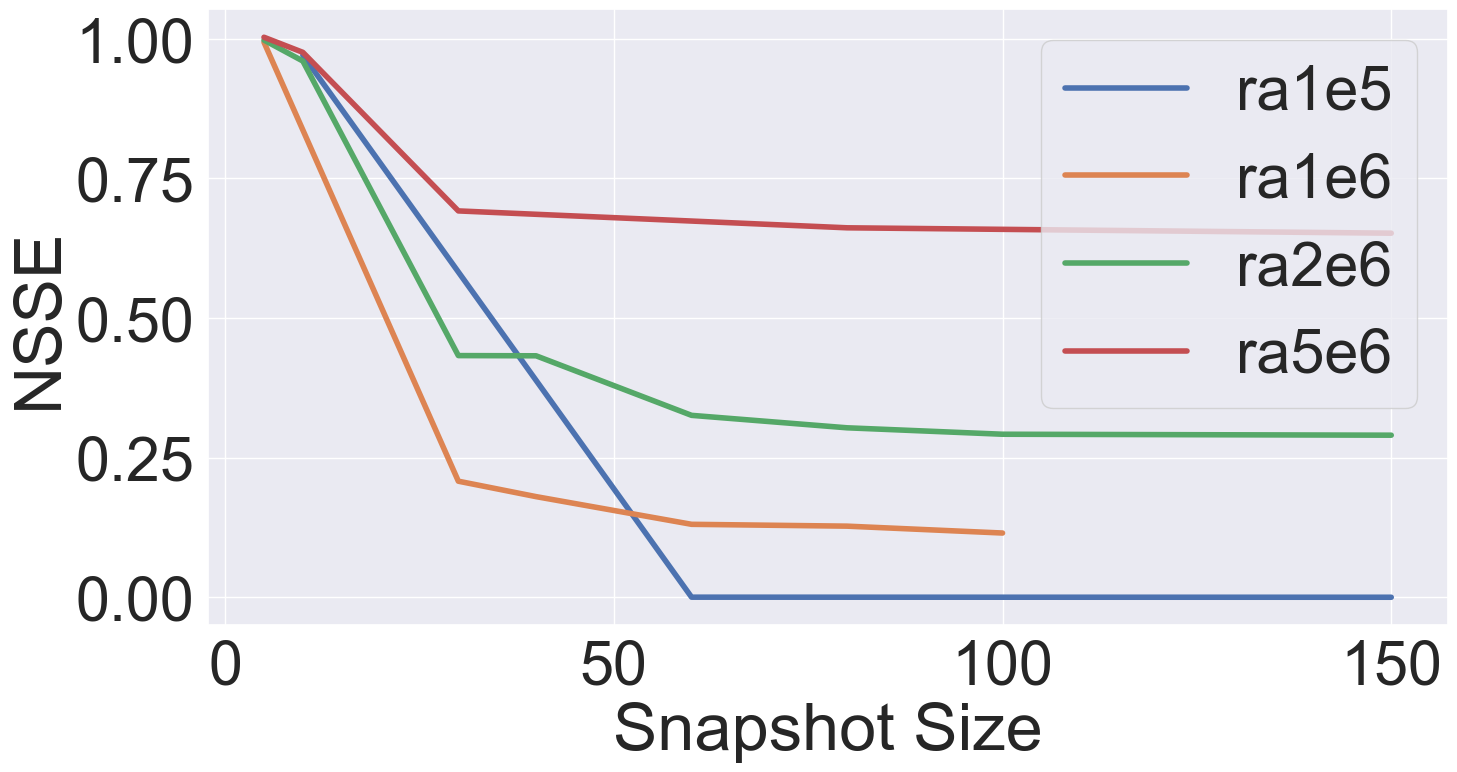}
        \caption{}
        \label{fig:results:sweep:kdmd}
    \end{subfigure}
    %\caption{ Correlation of hyperparameters to the NSSE on the test data. (a)-(c) shows picked parameters for all LRAN training runs in the sweep for $Ra=1e6$. (d) shows NSSE for tested KDMD snapshot size values grouped by $Ra$.}
    \caption{Correlation of hyperparameters to the NSSE on the test data. Panel (a) shows the correlation of LRAN sequence length and test NSSE. Each point represents 
    a training run in the LRAN sweep for $Ra=1e6$. Panel (b) shows the NSSE for tested KDMD snapshot size values grouped by $Ra$.}
    \label{fig:results:sweep:ra1e6}
\end{figure}
\begin{table}

\begin{subtable}{0.48\textwidth}
\centering
\begin{tabular}{||c | c c c||} 
 \hline
 Rayleigh Number & Sequence Length & Latent Dimension & $\delta$ \\ 
 \hline\hline
 1e5   & 18 & 200 & 0.9 \\
 1e6 & 20 & 400 & 0.9 \\
 2e6 & 20 & 400 & 0.9 \\
 5e6 & 25 & 500 & 0.9 \\
 \hline
\end{tabular}
\caption{LRAN hyperparameter}
\label{table:sweepresults:lran}
\end{subtable}

\vspace*{0.5 cm}

\begin{subtable}{0.48\textwidth}
\centering
\begin{tabular}{||c | c c||} 
 \hline
 Rayleigh Number & Snapshot Size & $\sigma$ \\ 
 \hline\hline
 1e5, 1e6, 2e6, 5e6   & 60 & 2 \\
 \hline
\end{tabular}
\caption{KDMD hyperparameter}
\label{table:sweepresults:kdmd}
\end{subtable}

\caption{Optimized parameters obtained from the hyperparameter searches in Experiment 1 and 2. (a) LRAN hyperparameters for each $Ra$. (b) KDMD hyperparameters valid for all $Ra$.}
\label{table:sweepresults}
\end{table}
Figure \ref{fig:results:sweep:ra1e6:seq} shows the most important hyperparameter for the LRAN while learning the RBC dynamics for Ra=1e6 - the length of the training sequences $\Tau$. There is a substantial decrease in the NSSE up to a sequence length of around 20. In comparison, the number of observables and the decaying weight delta are less relevant regarding the evaluation metric. The hidden loss had no significance in the search, therefore, we set $\beta=0$ for all following training runs. We also verified that the relationships of the hyperparameters in the search are similar for all Rayleigh numbers. Hence, we only show the curves for Ra=1e6. Based on the separate search results, a reasonably seeming configuration is picked for each Ra, shown in Table \ref{table:sweepresults:lran}.
The more turbulent settings benefit from a larger sequence length and latent dimension.

\subsection{Experiment 2}
The KDMD hyperparameter search results showed that a Gaussian kernel width of $\sigma = 2$ is the best choice. The effect of increasing the snapshot matrix size with a fixed sigma=2 is shown in Figure \ref{fig:results:sweep:kdmd}. As expected, the NSSE increases for increasing Ra but only negligibly improves for snapshot sizes larger than 60. The parameter values $\sigma = 2$ and a snapshot size equal to 60 were chosen for all Rayleigh numbers in the next experiment.

\subsection{Experiment 3}
\begin{figure*}
     \centering
     \begin{subfigure}[b]{0.35\textwidth}
         \centering
         \includegraphics[width=\textwidth]{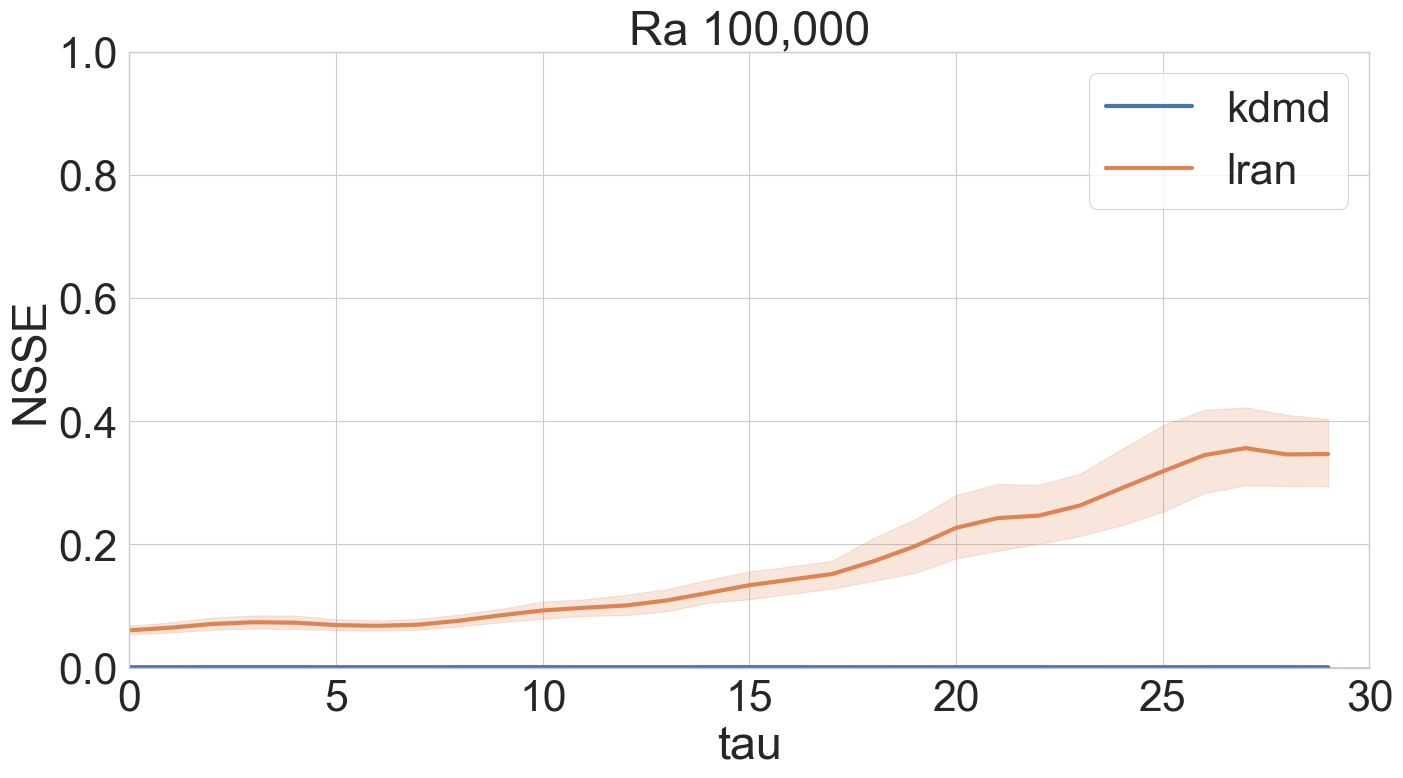}
         \caption{}
         \label{fig:resultsNSSE:ra1e5}
     \end{subfigure}
     \begin{subfigure}[b]{0.35\textwidth}
         \centering
         \includegraphics[width=\textwidth]{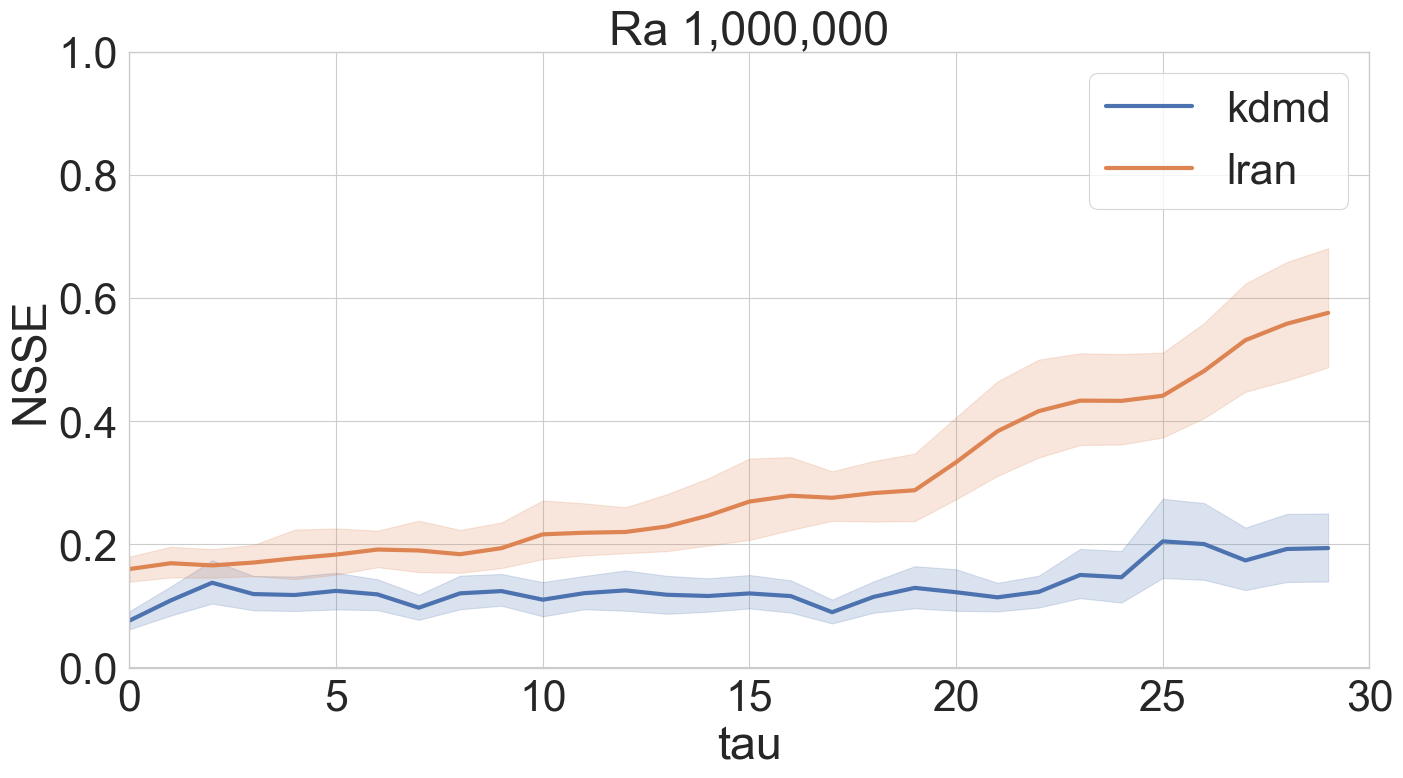}
         \caption{}
         \label{fig:resultsNSSE:ra1e6}
     \end{subfigure}
     \hfill
     \begin{subfigure}[b]{0.35\textwidth}
         \centering
         \includegraphics[width=\textwidth]{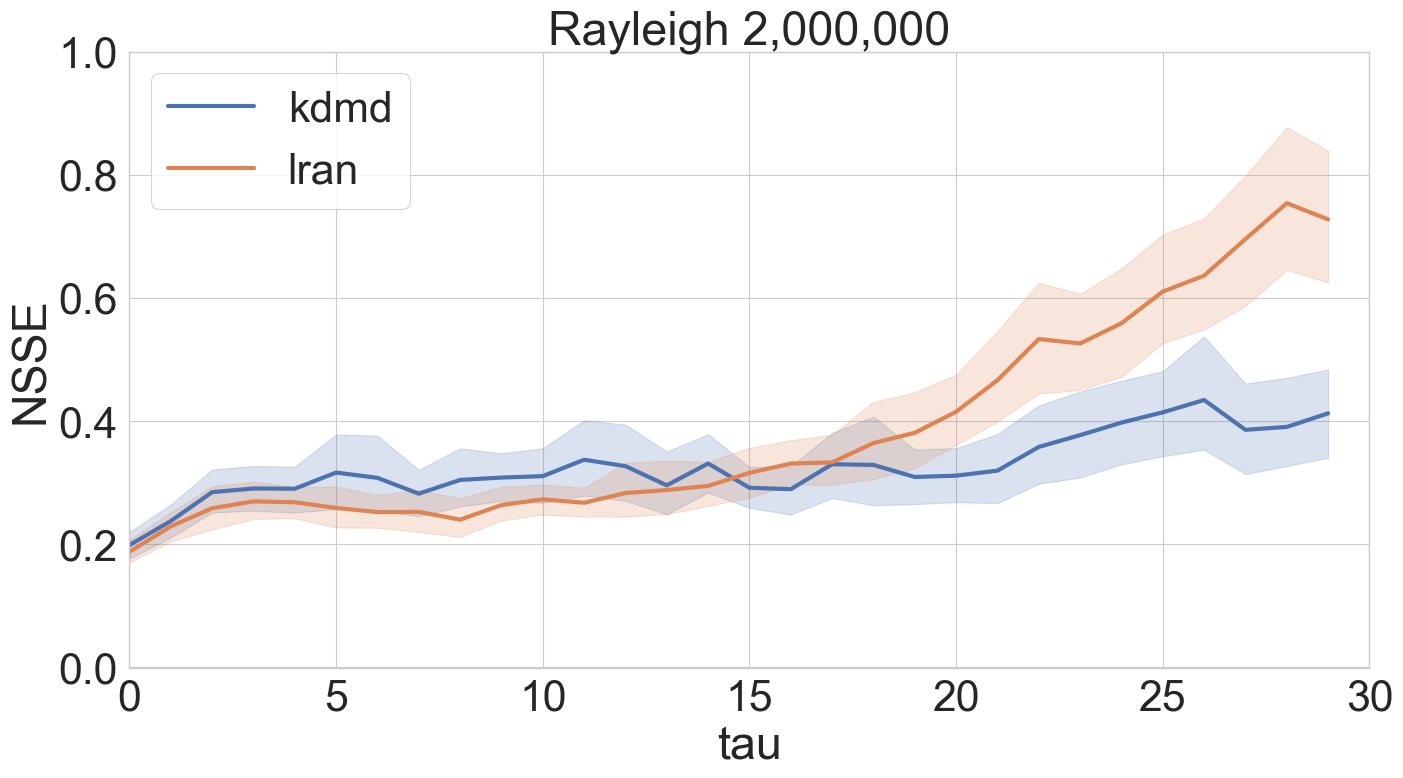}
         \caption{}
         \label{fig:resultsNSSE:ra2e6}
     \end{subfigure}
     \begin{subfigure}[b]{0.35\textwidth}
         \centering
         \includegraphics[width=\textwidth]{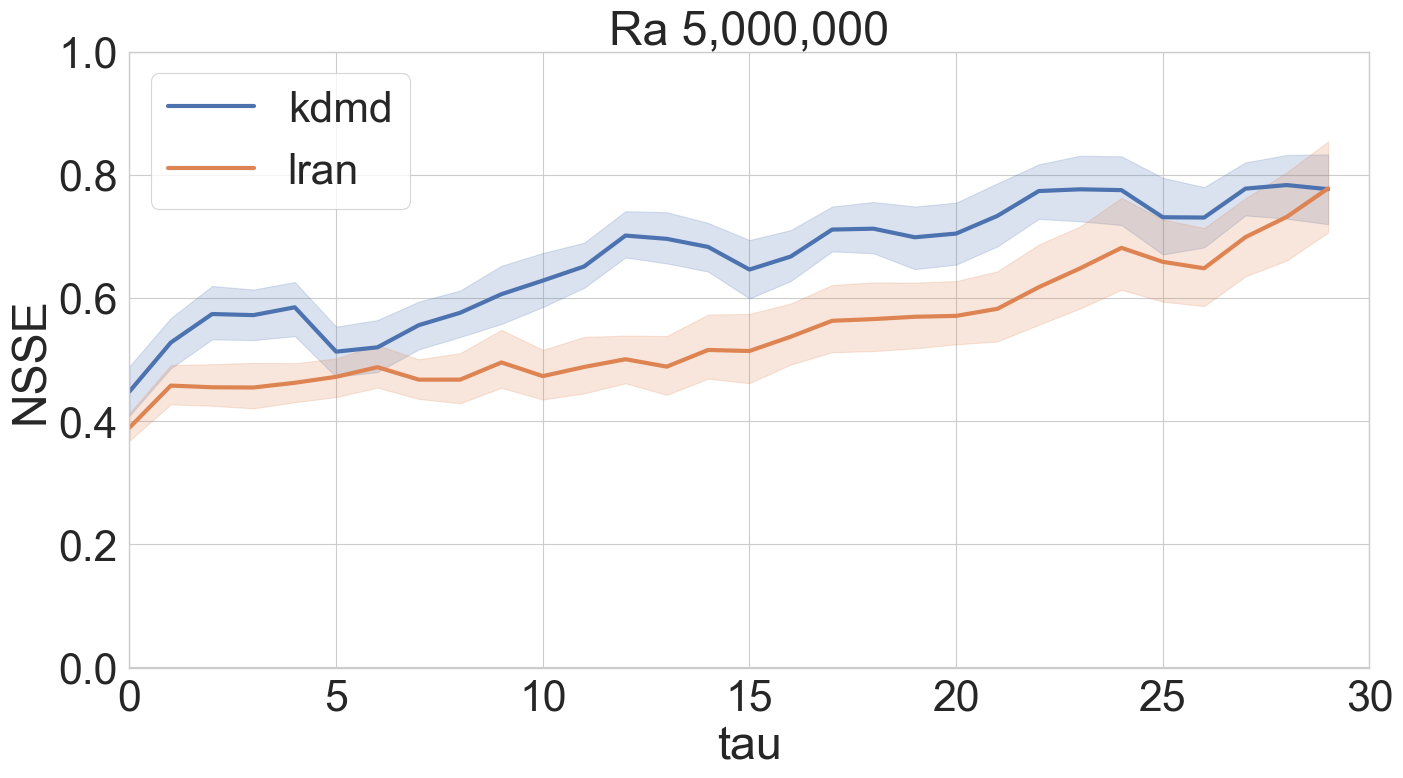}
         \caption{}
         \label{fig:resultsNSSE:ra5e6}
     \end{subfigure}
    \caption{The evolution of the NSSE in the test sequence averaged over all episodes. Shaded regions indicate the standard deviation.}
    \label{fig:resultsNSSE}
\end{figure*}
In Fig.~\ref{fig:resultsNSSE} we show the average test set accuracy of the LRAN and the KDMD predictions using the optimal hyperparameters for the different Rayleigh numbers. As shown in Fig.~\ref{fig:resultsNSSE:ra1e5}, for the lowest Rayleigh number $Ra=1e5$ the KDMD achieved nearly perfect accuracy ($\mathrm{NSSE} \approx 0)$ for the entire prediction window. The prediction errors of the LRAN were reasonably small but significantly higher than the KDMD error. Moreover, a sharp increase in error for timesteps $\tau > 10$ can be observed. Yet, upon visual inspection of the predicted convection fields made by the LRAN for this Rayleigh number (not shown here), we observed high qualitative agreement between the predictions and the ground truth 

At $Ra=1e6$, we see in Fig.~\ref{fig:resultsNSSE:ra1e6} that the relative growth of the prediction error for the KDMD compared to the case $Ra=1e5$ was large in comparison to the growth of the prediction error of the LRAN.

The differences observed in the relative growth of the prediction error with respect to the Rayleigh number are indeed a trend that continued with increasing Rayleigh number: For $Ra=2e6$, the relative error increase compared to $Ra=1e6$ was again substantially higher for the KDMD than for the LRAN.
For the first half of the prediction window, the KDMD and LRAN showed similar prediction errors. As before, the LRAN prediction errors sharply increased at more distant prediction times from the entry point.

% Examples
\begin{figure*}
    \centering
    \begin{subfigure}[b]{0.3\textwidth}
        \centering
        \includegraphics[width=\textwidth]{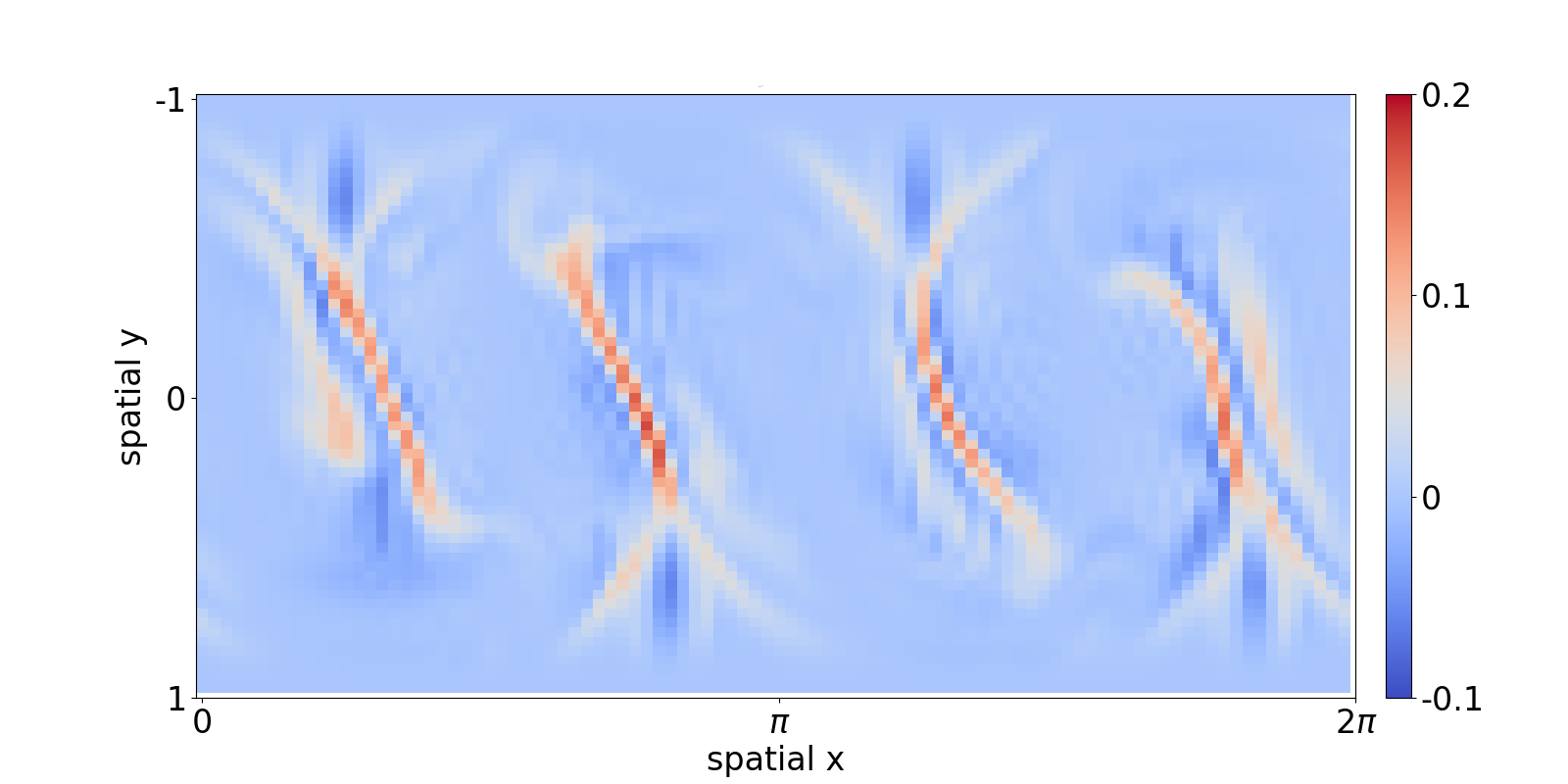}
        \caption{Ground Truth at $\tau=1$}
        \label{fig:results:ra5e6:gt1}
    \end{subfigure}
    \begin{subfigure}[b]{0.3\textwidth}
        \centering
        \includegraphics[width=\textwidth]{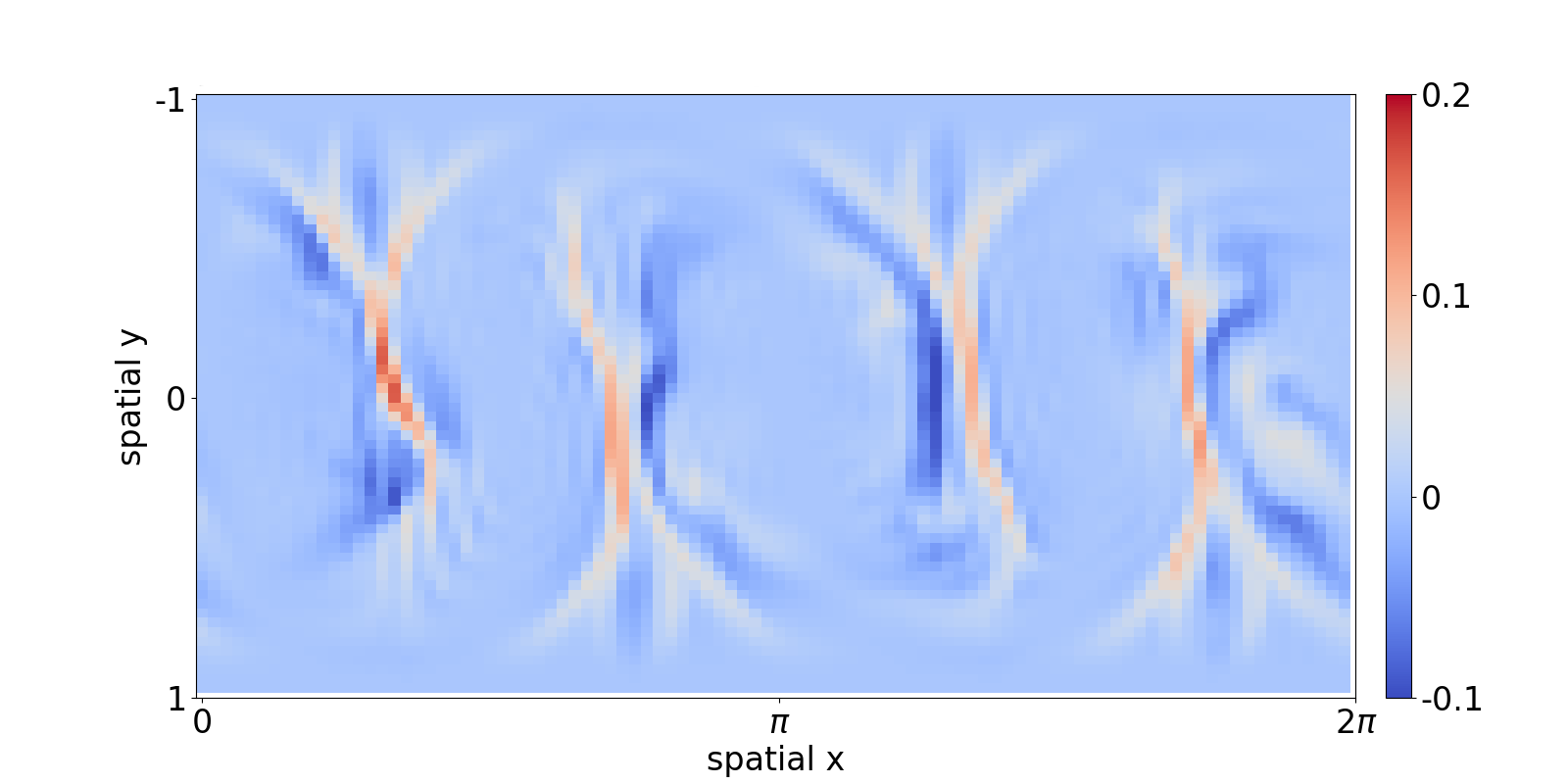}
        \caption{Ground Truth at $\tau=10$}
        \label{fig:results:ra5e6:gt10}
    \end{subfigure}
    \begin{subfigure}[b]{0.3\textwidth}
        \centering
        \includegraphics[width=\textwidth]{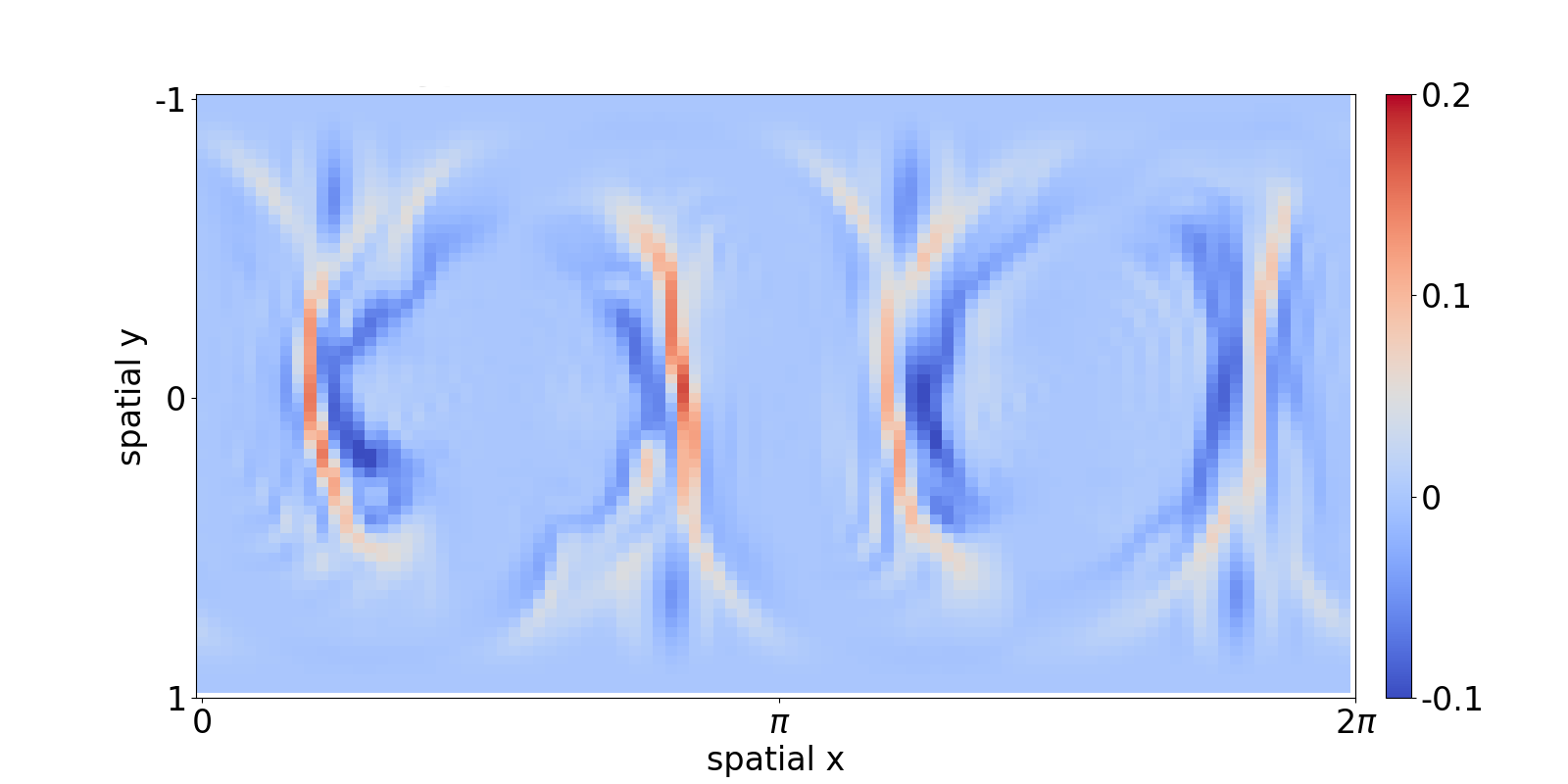}
        \caption{Ground Truth at $\tau=25$}
        \label{fig:results:ra5e6:gt25}
    \end{subfigure}
     
    \begin{subfigure}[b]{0.3\textwidth}
        \centering
        \includegraphics[width=\textwidth]{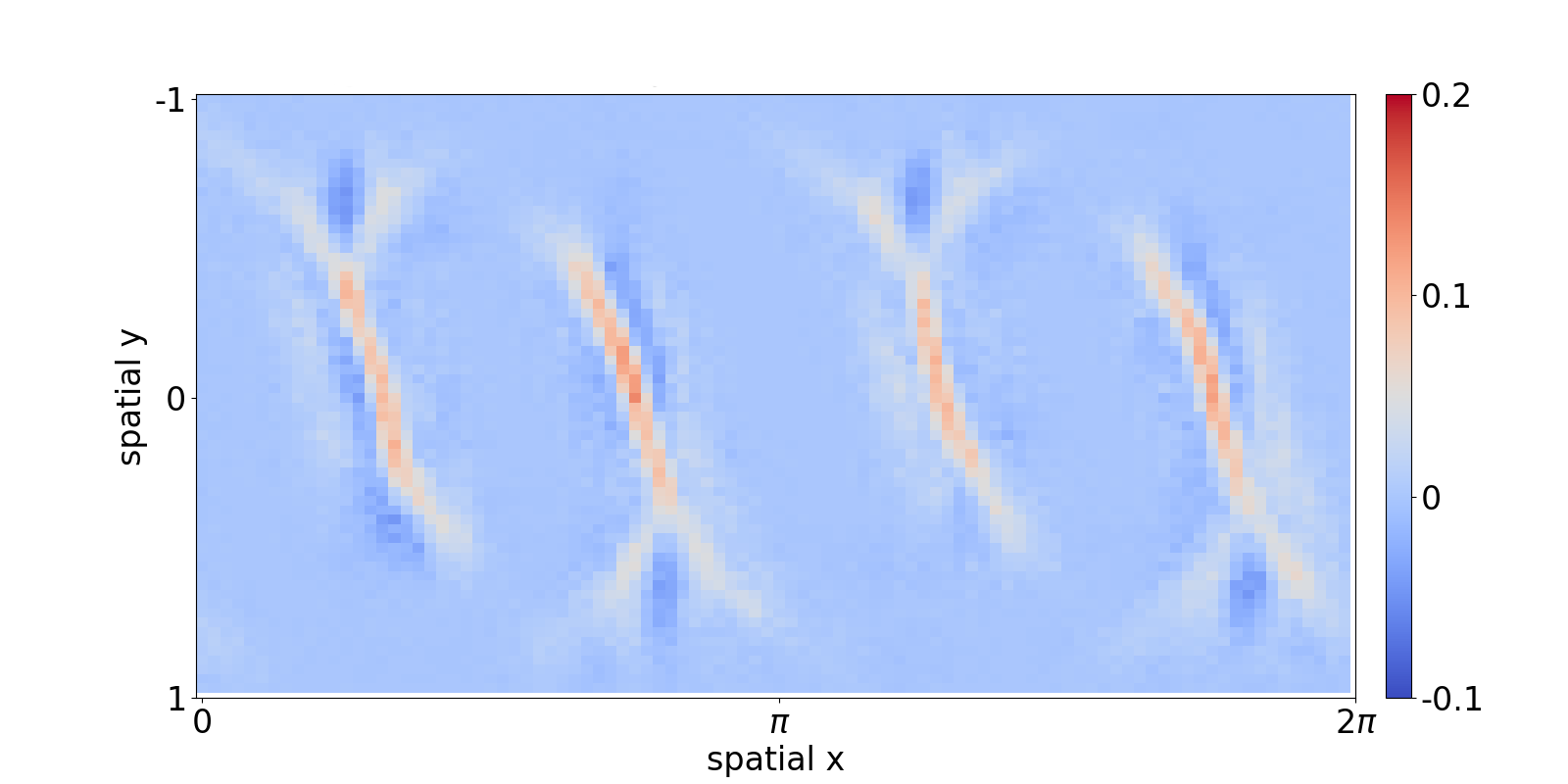}
        \caption{Prediction at $\tau=1$}
        \label{fig:results:ra5e6:p1}
    \end{subfigure}
    \begin{subfigure}[b]{0.3\textwidth}
        \centering
        \includegraphics[width=\textwidth]{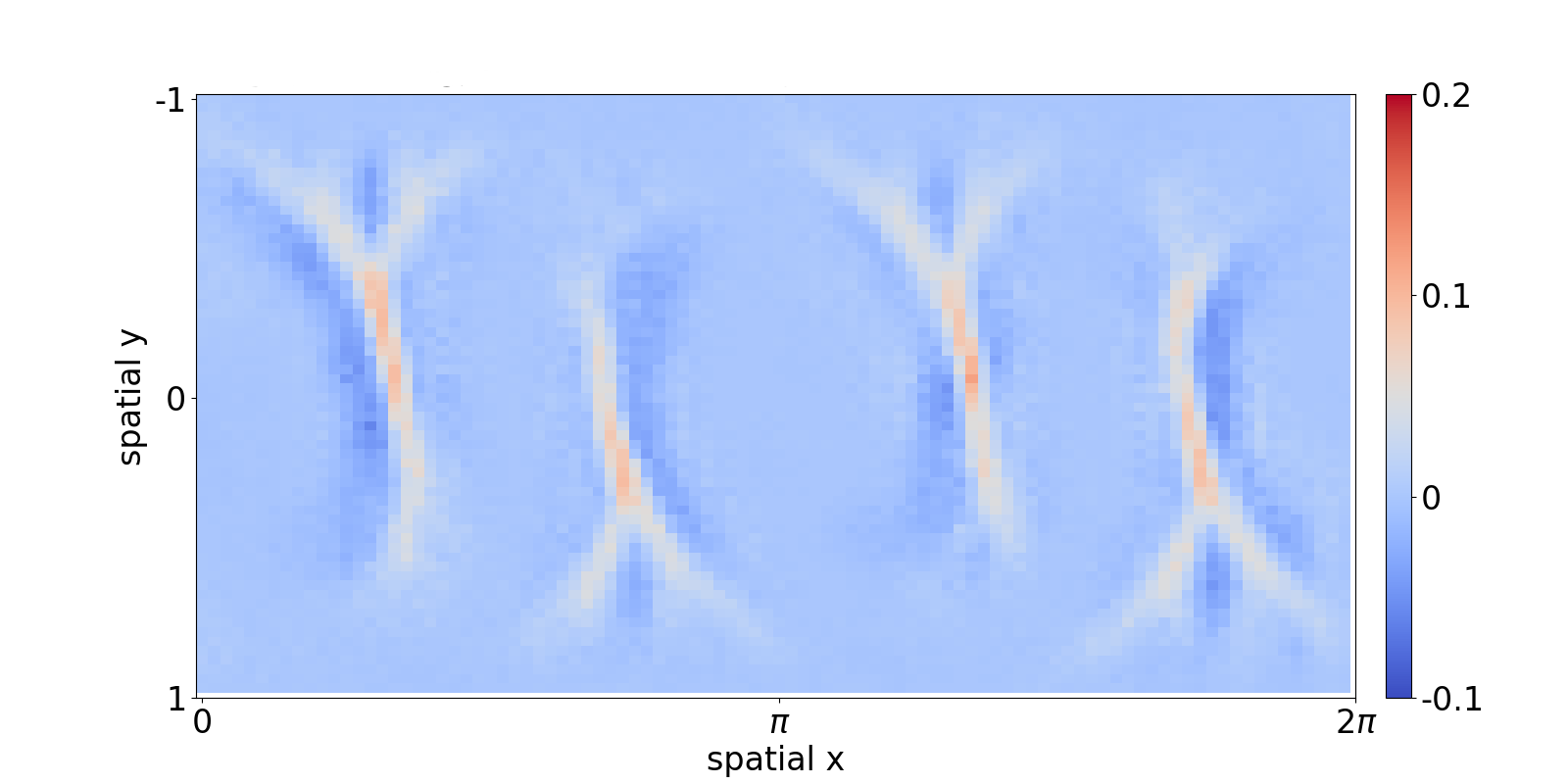}
        \caption{Prediction at $\tau=10$}
        \label{fig:results:ra5e6:p10}
    \end{subfigure}
    \begin{subfigure}[b]{0.3\textwidth}
        \centering
        \includegraphics[width=\textwidth]{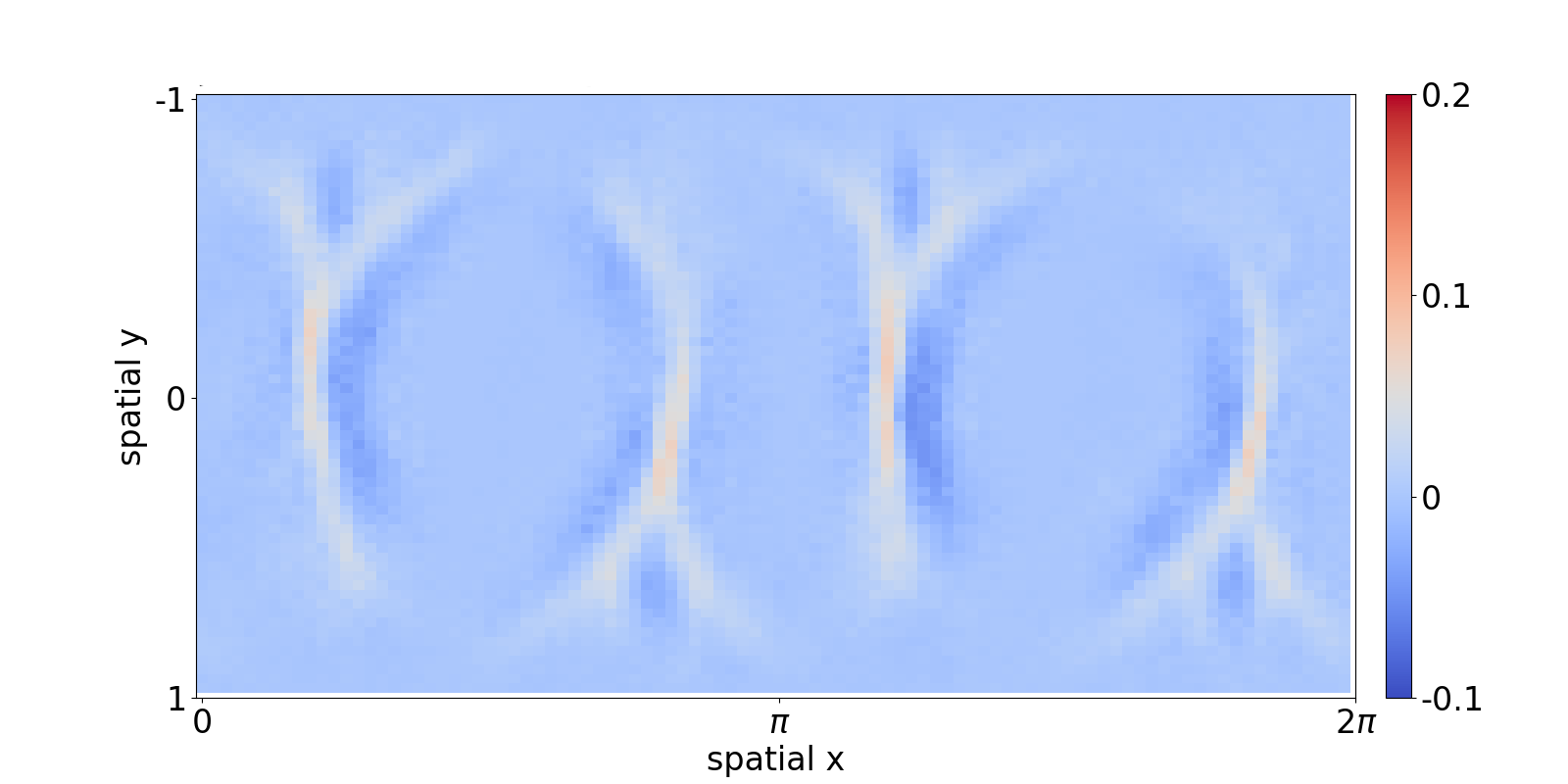}
        \caption{Prediction at $\tau=25$}
        \label{fig:results:ra5e6:p25}
    \end{subfigure}
    \caption{Examples for ground truth and LRAN predicted local convective fields for $Ra=5e6$.} 
    \label{fig:results:ra5e6}
\end{figure*}

\begin{figure}
    \centering
    \includegraphics[width=0.48\textwidth]{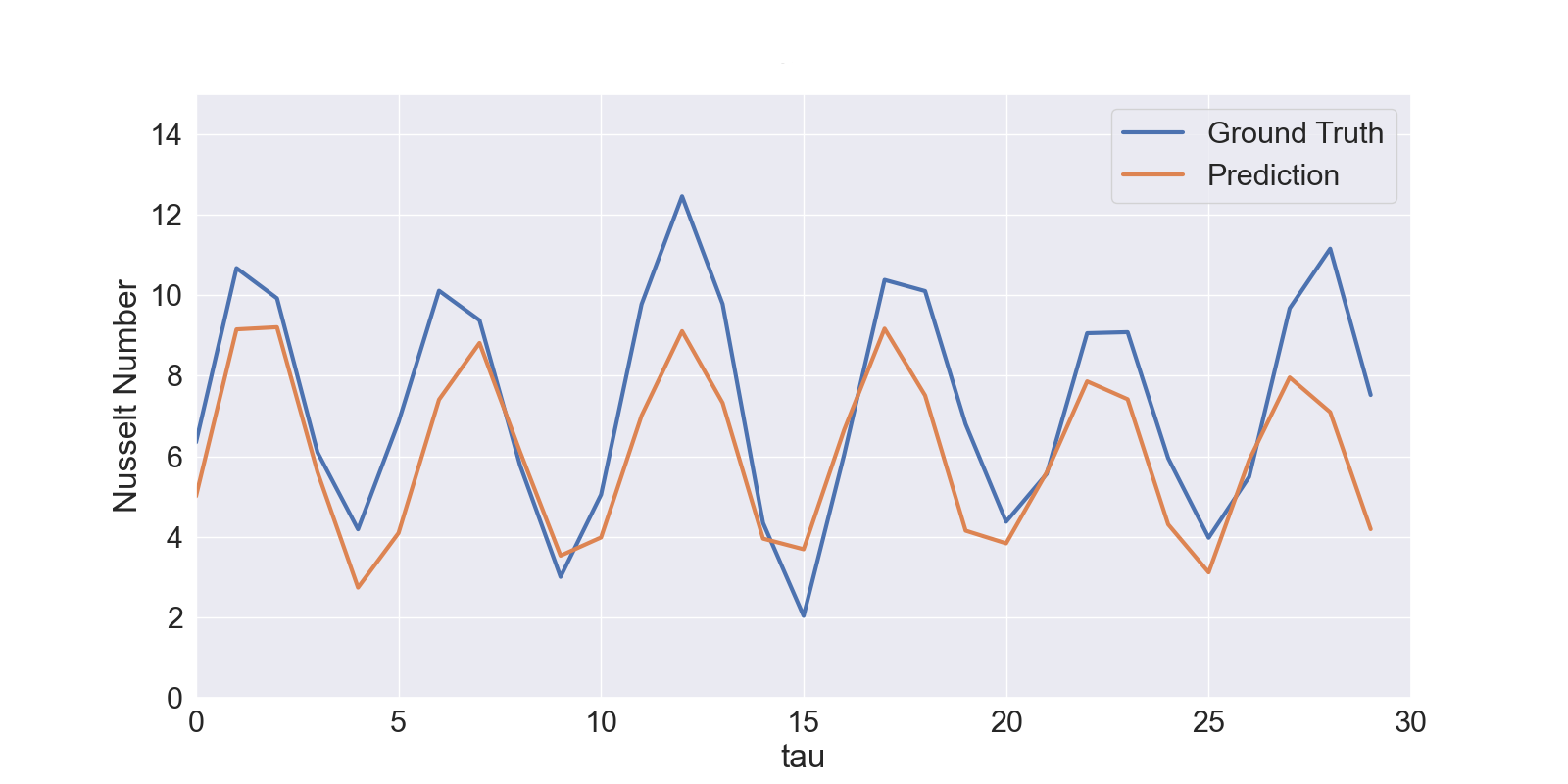}
    \caption{Original and LRAN prediction for the Nusselt Number of Episode 12 for $Ra=5e6$}
    \label{fig:results:ra5e6:nusselt}
\end{figure}

Because of the slower growth in prediction error of the LRAN for increasing turbulence, we observed that
for the most turbulent case at $Ra=5e6$ (Fig.~\ref{fig:resultsNSSE:ra5e6}) the LRAN consistently achieved lower error than the KDMD over the entire prediction window.
Fig.~\ref{fig:results:ra5e6} shows ground truth fields and the predictions made by the LRAN at a selection of time lags from the entry point. Larger structures were predicted reasonably well. The general structure of the convection field was still predicted correctly at large prediction times, although the field was fainter.
Fig.~\ref{fig:results:ra5e6:nusselt} shows the Nusselt number computed for both the LRAN predicted fields and the ground truth fields. We observe high agreement in the first half of the prediction window. In the second half of the window, the times of the peaks corresponded, but the amplitudes differed significantly for some fields.

\section{Discussion} \label{sec:discussion}
The nature of the convection and the amount of turbulence determines the effectiveness of the LRAN architecture.

For the lowest Rayleigh number we considered ($Ra=1e5$), the convective flow structures have repetitive behavior over time. Although the LRAN predictions achieved high qualitative agreement with the ground truth, the KDMD error was much lower.
This can be explained by the \ac{DMD} formulation in Eq.~\eqref{eq:DMDFormulation}, which allows for the representation of periodic dynamical structures.
Hence, the KDMD could capture these repetitive convection patterns almost perfectly, which explains why it achieved near-optimal prediction error in this case.
We trained the LRAN on sequence lengths of at most 25 to learn the underlying evolution of the convection.
Therefore, the LRAN could not capture long periodic patterns as well as the KDMD, which led to worse performance.

The flow becomes more turbulent and less repetitive for higher Rayleigh numbers. Therefore, a more complex observable function space is necessary to represent the faster turbulent dynamics. The KDMD method's observable function space is fixed and is restricted by the choice and parameters of the kernel, which is not a flexible approach to capture the more complex dynamics. In this case, the KDMD can only represent the global periodic patterns and not the turbulent features of the flow. Hence, the KDMD's error compared to the ground truth increased for higher Rayleigh numbers. In contrast, the LRAN has greater flexibility in learning an appropriate observable function space. The LRAN architecture allows for increasing the number of observables required to capture the higher velocity turbulent flows and non-repetitive patterns, which occur at higher Rayleigh numbers. Moreover, the encoder can learn highly complex observables from the data. Therefore, the flexibility of the LRAN in learning a large number of complex observables from data becomes increasingly relevant for larger Rayleigh numbers. We conjecture that this higher flexibility allows the LRAN to capture non-periodic and turbulent patterns more accurately than KDMD.

In the examples shown in Fig.~\ref{fig:results:ra5e6}, the current version of the LRAN for $Ra=5e6$ captures the non-periodic nature of the dynamics of the larger convective flows correctly. However, it does not include the smaller details of the flow. This is because the current autoencoder architecture is not complex enough to represent these smaller details, which we verified by studying the reconstructions of the autoencoder alone.
If the application requires more fine-grained predictions, it should be possible to improve the result by increasing the complexity of the autoencoder architecture and the number of observables. This requires much more data to ensure that the large architecture learns dynamics that generalize.

\section{Conclusion} \label{sec:conclusion}
We systematically studied the features and performance of a Koopman-based Machine Learning architecture called LRAN for learning RBC dynamics and compared the performance to the more traditional KDMD method.
Our study showed better prediction accuracy of the LRAN architecture for more turbulent scenarios than KDMD predictions.
Specifically, the LRAN architectures we trained showed to be appropriate for dealing with complex flows in settings where accurate short-time predictions are crucial.

We plan to extend the work in several directions: Besides studying the method in changing environments and further improving the loss function, we aim to study and improve the latent space by incorporating physical knowledge about the system. For instance, incorporating symmetries and invariances in the Machine Learning formulation would be interesting and could potentially accelerate training.\\
An important future direction is the Koopman-based modeling of controlled RBC dynamics in which small temperature deviations can be dynamically applied at the bottom layer to steer the convection in the system, which is relevant to various industrial processes. A strong feature of the linear Koopman model is the possibility of using model-based control techniques for computing the actuation of the system that reaches a desired state. Potentially, this could be a more accurate and data-efficient methodology than current model-free approaches used in this domain.

% For papers published in translation journals, please give the English 
% citation first, followed by the original foreign-language citation \cite{b6}.

\bibliographystyle{IEEEtran}
\bibliography{IEEEabrv,bibliography}

\begin{appendices}
\section{Source Code and Data}
The source code used for methods, data, and hyperparameter tuning is available on GitHub at https://github.com/SAIL-project/RayleighBenard. The simulation data can be requested from the first author upon reasonable request.

\section{Convolutional Autoencoder}
\label{app:cae}
\begin{table}[h]
\centering
\begin{tabular}{|c |c || c |c |} 
 \hline
 \multicolumn{2}{|c||}{Encoder} &
 \multicolumn{2}{|c|}{Decoder} \\
 \hline
 Layer Type & Output Shape & Layer Type & Output Shape  \\ 
 \hline
 Input & [1, 1, 64, 96]   & Input & [1, Latent Dim] \\
 Conv2D & [1, 32, 32, 48] & Dense & [1, 12288] \\
 Conv2D & [1, 64, 32, 48] & Deconv2D & [1, 32, 16, 24] \\
 Conv2D & [1, 32, 16, 24] & Deconv2D & [1, 64, 32, 48] \\
 Conv2D & [1, 32, 32, 24] & Deconv2D & [1, 32, 32, 48] \\
 Flatten & [1, 12288]     & Deconv2D & [1, 1, 64, 96] \\
 Dense & [1, Latent Dim]  & & \\
 \hline
\end{tabular}

\caption{Convolutional Autoencoder Architecture}
\label{table:ae}
\end{table}
\end{appendices}

\end{document}